\documentclass[letterpaper]{article} 
\usepackage{aaai2027}  
\usepackage[hyphens]{url}  
\usepackage{graphicx} 
\usepackage{natbib}  
\usepackage{caption} 
\usepackage{algorithm}
\usepackage{algorithmic}

\usepackage{newfloat}
\usepackage{multirow}
\usepackage{amsfonts}
\usepackage{amsmath}
\usepackage{listings}
\DeclareCaptionStyle{ruled}{labelfont=normalfont,labelsep=colon,strut=off} 
\floatstyle{ruled}
\newfloat{listing}{tb}{lst}{}
\floatname{listing}{Listing}

\usepackage{booktabs}

\title{Frequency-Domain Dual-Branch Fusion for Medical Visual Question Answering}
\author{
    Yusra Tariq, Rakesh Chandra Joshi
}
\affiliations{
Department of Artificial Intelligence, Amity University, Noida, India
    
}

\begin{document}

\maketitle

\begin{abstract}
Medical Visual Question Answering (VQA) requires aligning subtle visual evidence, including lesion texture, boundary sharpness, and diffuse density changes, with clinical language. Existing multimodal fusion approaches operating in the spatial domain may not fully exploit complementary frequency information present in visual and textual representations. We introduce a dual-branch frequency-domain fusion module that conditions spectral filtering on the input question, enabling adaptive selection of global low-frequency structure and fine-grained high-frequency detail before reconstructing the spatial representation for answer generation. To provide a richer spectrum for filtering, we extract complementary features from early texture-sensitive and final semantic layers of a frozen BiomedCLIP encoder and align both with the question representation using a symmetric InfoNCE objective prior to staged joint training with a BioBART decoder. We pretrain the proposed model on PMC-VQA and fine-tune it on the VQA-RAD and SLAKE benchmarks, demonstrating that frequency-aware multimodal fusion improves medical VQA performance while maintaining a lightweight and efficient architecture.
\end{abstract}

\section{Introduction}

Medical Visual Question Answering (Med-VQA) requires a model to answer
clinical questions by grounding them in fine grained visual evidence from
medical images. Unlike general visual question answering, where object
recognition is often sufficient, Med-VQA demands the interpretation of
subtle anatomical structures, lesion boundaries, tissue texture, and
diffuse pathological changes. Consequently, the mechanism that fuses visual
and textual representations plays a central role in determining the quality
of multimodal reasoning. A fusion strategy that treats all spatial regions
and feature scales uniformly may fail to distinguish subtle textural
abnormalities from larger structural patterns that are equally important
for clinical decision making.

Recent Med-VQA approaches primarily improve performance by combining biomedical vision-language pretraining with spatial-domain cross-attention fusion, where question tokens attend over image patch embeddings. While effective at aligning semantic content, this paradigm does not separate global structure from fine-grained texture before fusion, leaving both entangled in the same representation, and incurs quadratic computational cost as token sequences grow.

Frequency-domain learning offers an alternative to spatial fusion: transforming representations via the Fourier transform decomposes them into globally-aggregated low-frequency structure and high-frequency texture. Global Filter Networks showed that learnable filtering in the frequency domain is an efficient alternative to spatial token mixing for visual representation learning \citep{rao2021gfnet}, and FSRU extended this to multimodal rumor detection via unimodal spectrum compression and cross-modal spectrum co-selection \citep{lao2024fsru}. However, frequency-domain fusion remains unexplored for Medical VQA, where fine-grained image features must be integrated with clinically meaningful language representations.

Motivated by these observations, we propose a dual branch frequency domain
fusion framework for Medical Visual Question Answering. Rather than
directly fusing spatial representations, our framework transforms visual
and textual features into the frequency domain, performs question guided
spectral filtering and cross modal interaction, and reconstructs enhanced
spatial representations for answer generation. The image branch operates on
two dimensional Fourier representations extracted from a frozen
BiomedCLIP vision encoder, while the text branch performs one dimensional
Fourier analysis over contextual question embeddings. A bank of learnable
spectral filters performs unimodal spectrum compression, followed by a
cross modal gating mechanism that selectively emphasizes or suppresses
frequency components according to complementary information from the other
modality. This design enables the model to adaptively balance global
structural information and fine grained texture based on the clinical
question being answered.

We pretrain the proposed model on PMC-VQA~\citep{zhang2023pmcvqa} and evaluate it on the VQA-RAD and SLAKE benchmarks using Exact Match and Token F1 for both open-ended and closed-ended questions. The experimental results demonstrate the effectiveness of the proposed frequency-domain fusion strategy for multimodal medical reasoning.

Our contributions are threefold.

\begin{itemize}
\item We propose a dual branch question conditioned frequency domain fusion
module for Medical Visual Question Answering that performs spectral
filtering and cross modal gating before reconstructing enhanced spatial
representations.

\item We introduce a dual depth visual representation strategy that combines
early texture sensitive features with late semantic features from a frozen
BiomedCLIP encoder and aligns them with question representations through a
contrastive pretraining objective.

\item We demonstrate the effectiveness of the proposed frequency-domain fusion framework through extensive experiments on the VQA-RAD~\citep{lau2018vqarad} and SLAKE~\citep{liu2021slake} benchmarks after pretraining on PMC-VQA.
\end{itemize}

\section{Related Work}

Early Med-VQA methods combined CNN visual encoders with recurrent question encoders, fusing modalities via concatenation or bilinear pooling; these established task feasibility but were limited by local receptive fields and scarce annotated medical data. Recent approaches instead pair biomedical vision-language pretraining, such as BiomedCLIP \citep{zhang2023biomedclip}, with spatial-domain cross-attention fusion, improving semantic alignment but entangling global structure and local texture within the same fused representation, at quadratic cost in token count.

Frequency-domain learning offers an efficient alternative to spatial or sequential token mixing: GFNet \citep{rao2021gfnet} and FNet \citep{leethorp2022fnet} replace spatial self-attention with learnable filtering in the 2D and 1D Fourier domains, respectively, for unimodal vision and language representation. FSRU \citep{lao2024fsru} extends this to multimodal fusion via unimodal spectrum compression and cross-modal spectrum co-selection, though for rumor detection rather than question-conditioned reasoning. Frequency-domain fusion for Med-VQA remains unexplored: existing methods rely on spatial-domain cross-attention \citep{zhang2023biomedclip}, motivating the question-conditioned frequency-domain fusion approach developed here.

\subsection{Problem Definition}

We formulate Medical Visual Question Answering (Med-VQA) as a multimodal question answering task. Let $\mathcal{D} = \{(v_i, q_i, a_i, t_i)\}_{i=1}^{N}$ denote a Med-VQA dataset of $N$ samples, where $v_i$ is a medical image, $q_i$ is the associated clinical question represented as a token sequence, $a_i$ is the corresponding ground-truth answer, also represented as a token sequence, drawn from a shared vocabulary $\mathcal{V}$ with $q_i$, and $t_i \in \{\text{yes/no}, \text{open}, \text{other}\}$ is the answer-type label associated with the sample. Given an image-question pair $(v, q)$, our goal is to learn a parametric mapping

\begin{equation}
f_\theta : (v, q) \longmapsto \hat{a},
\end{equation}

where $\theta$ denotes the learnable parameters of the model and $\hat{a}$ is the predicted answer sequence generated by jointly reasoning over the visual content of $v$ and the linguistic content of $q$. The objective is to learn the parameters $\theta$ such that the predicted answer $\hat{a} = f_\theta(v, q)$ closely matches the ground-truth answer $a$, enabling clinically accurate and contextually relevant responses to medical image-question pairs.

During training, the model additionally predicts the answer type $\hat{t}$ from a shared encoder representation, so that Equation (1) is optimized jointly with an auxiliary classification objective over $t$. This auxiliary objective encourages the encoder to represent coarse answer-type structure alongside fine-grained answer content, and is used only as a training-time regularizer. At inference time, only $\hat{a} = f_\theta(v, q)$ is used for evaluation against the ground-truth answer $a$.

\section{Methodology}

\subsection{Overall Framework}

\begin{figure*}[t]
    \centering
    \includegraphics[width=0.95\linewidth]{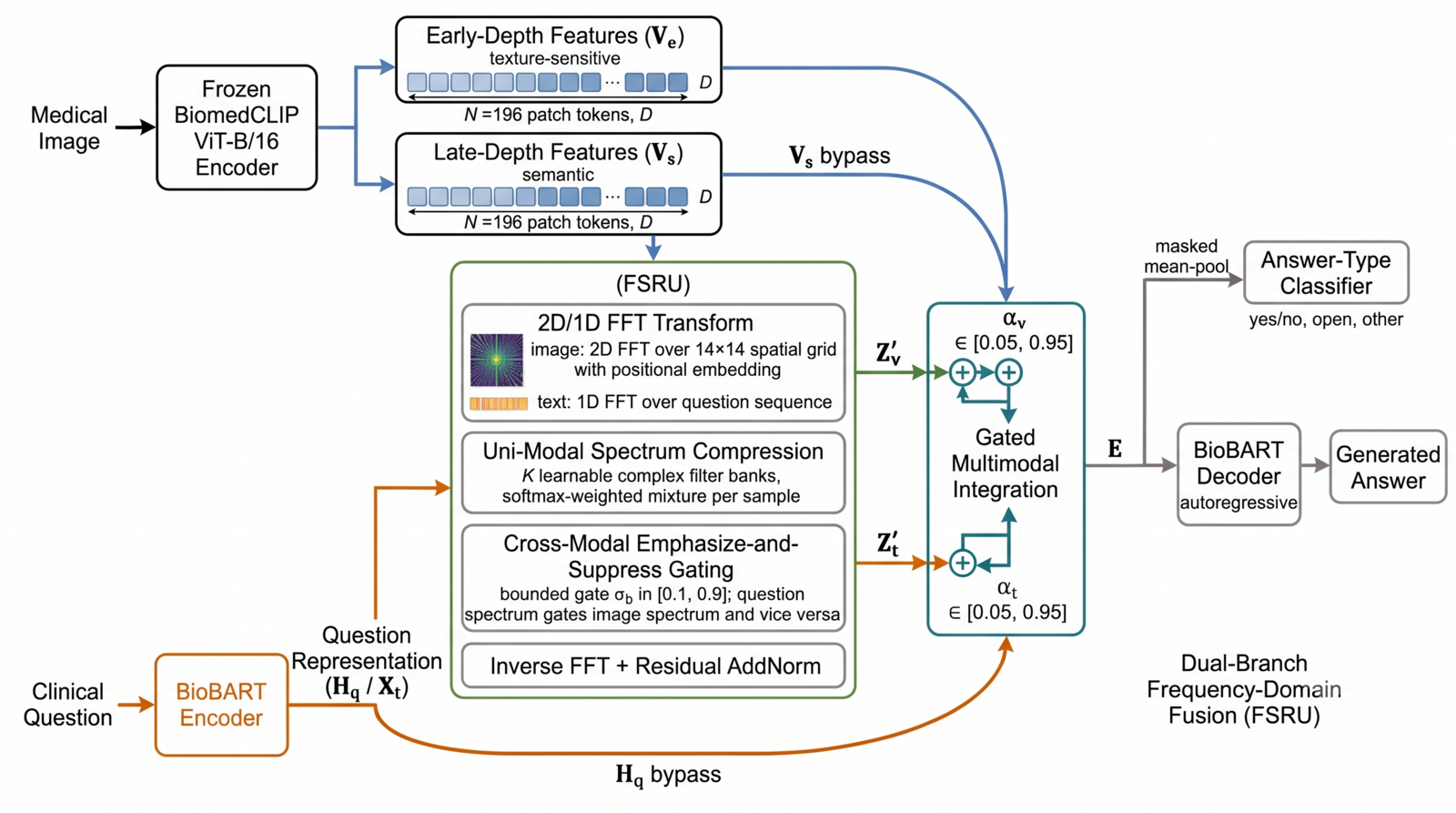}
    \caption{Overall architecture. A frozen BiomedCLIP encoder extracts early- and late-depth visual features ($\mathbf{V}_e$, $\mathbf{V}_s$); a BioBART encoder produces the question representation ($\mathbf{H}_q$). $\mathbf{V}_e$ and the question representation are processed by the dual-branch frequency-domain fusion module (FSRU), while $\mathbf{V}_s$ and $\mathbf{H}_q$ bypass FSRU directly. Gated multimodal integration combines both paths into the fused representation $\mathbf{E}$, which feeds an auxiliary answer-type classifier and the BioBART decoder for answer generation.}
    \label{fig:architecture}
\end{figure*}

Given an image-question pair $(v, q)$, our model predicts the answer $\hat{a}$ through four stages: (1) a frozen vision transformer extracts patch-level visual features at two depths, capturing low-level spatial detail and high-level semantic content respectively; (2) a pretrained sequence-to-sequence text encoder produces contextual question representations; (3) a dual-branch frequency-domain fusion module filters the low-level visual and question representations in the spectral domain and exchanges information between them via learned cross-modal gates; and (4) the frequency-enhanced representations are additively fused with the late-semantic visual stream and the original textual stream, and passed to a sequence decoder that autoregressively generates the answer. An auxiliary classification head jointly predicts the coarse answer category (yes/no, open, other), providing an additional training signal without altering the generation pathway. Figure~\ref{fig:architecture} illustrates the overall pipeline.

\subsection{Visual and Question Feature Extraction}

\paragraph{Dual-depth visual encoding.} A frozen Vision Transformer $\Phi$ (BiomedCLIP~\citep{zhang2023biomedclip}), pretrained on biomedical image-text pairs, decomposes an image $v$ into $N=196$ patch tokens ($14 \times 14$ grid). We extract hidden states at two depths: an \emph{early} intermediate block, $\mathbf{T}_e \in \mathbb{R}^{N \times d}$, retaining local texture and edge information relevant to radiological findings, and the \emph{final} block, $\mathbf{T}_s \in \mathbb{R}^{N \times d}$, carrying globally contextualized semantics. Both are linearly projected into the shared model dimension $D$:
\begin{equation}
\mathbf{V}_e = \mathbf{T}_e \mathbf{W}_e + \mathbf{b}_e, \qquad
\mathbf{V}_s = \mathbf{T}_s \mathbf{W}_s + \mathbf{b}_s.
\end{equation}
$\mathbf{V}_e$ feeds the frequency-domain fusion module (Section 3.3); $\mathbf{V}_s$ is later fused additively (Section 3.4). $\Phi$ remains frozen throughout training; only $\mathbf{W}_e$ and $\mathbf{W}_s$ are learned.

\paragraph{Question encoding.} The question $q$ is encoded by the encoder of BioBART \citep{yuan2022biobart}, a pretrained biomedical sequence-to-sequence language model,
\begin{equation}
\mathbf{H}_q = \mathrm{Encoder}(q) \in \mathbb{R}^{L_q \times D},
\end{equation}
and linearly projected to $\mathbf{X}_t = \mathbf{H}_q \mathbf{W}_t + \mathbf{b}_t \in \mathbb{R}^{L_q \times D}$, the representation consumed by the fusion module. Unlike $\Phi$, the question encoder is fine-tuned jointly in later training stages.
\paragraph{Contrastive feature alignment (pretraining).} Prior to joint training, the early- and late-depth visual projections are aligned with the question representation using a symmetric InfoNCE objective~\citep{oord2018cpc}, following the temperature-scaled contrastive formulation of CLIP~\citep{radford2021clip}. Mean-pooled, $\ell_2$-normalized projections of $\mathbf{V}_e$ and $\mathbf{V}_s$ are each contrasted against the pooled question representation $\mathbf{X}_t$ across the batch, with a learnable temperature (logit scale) shared across both depths. During this stage, only the projection layers ($\mathbf{W}_e$, $\mathbf{W}_s$, $\mathbf{W}_t$) and the temperature parameter are updated; all other parameters, including the frozen vision backbone and the question encoder itself, remain fixed. This warmup stage ensures both visual depths enter the frequency-domain fusion module (Section 3.3) already loosely aligned with the question's semantic space, before any generation-based supervision is introduced.

\subsection{Dual-Branch Frequency-Domain Fusion}

The core of our approach filters and cross-modulates $\mathbf{V}_e$ and $\mathbf{X}_t$ in the frequency domain, treating the image and text streams as independent spectral branches connected only through learned cross-modal gates.

\paragraph{Spectral transform.} The early-visual tokens are reshaped to their native $14 \times 14$ grid and combined with a learned positional embedding $\mathbf{P} \in \mathbb{R}^{14 \times 14 \times D}$, compensating for the translation equivariance of Fourier-domain filtering, which would otherwise treat the spatial grid as circular (toroidal).
\begin{equation}
\mathbf{G} = \mathrm{reshape}(\mathbf{V}_e) + \mathbf{P}.
\end{equation}
A two-dimensional real FFT is applied over the spatial dimensions and a one-dimensional real FFT over the question sequence:
\begin{align}
\mathbf{F}_v &= \mathcal{F}_{2D}(\mathbf{G}) \in \mathbb{C}^{14 \times 8 \times D}, \\
\mathbf{F}_t &= \mathcal{F}_{1D}(\mathbf{X}_t) \in \mathbb{C}^{L_t' \times D},
\end{align}
where $L_t' = \lfloor L_q/2 \rfloor + 1$.

\paragraph{Uni-modal spectrum compression.} Each branch is filtered by a bank of $K$ learnable complex filters, combined with sample-dependent mixture weights derived from the pooled magnitude spectrum:
\begin{align}
w^{v} &= \mathrm{softmax}\!\left(g_v\!\left(\textstyle\frac{1}{HW}\sum |\mathbf{F}_v|\right)\right), \\
\tilde{\mathbf{F}}_v &= \sum_{k=1}^{K} w^{v}_k \left(\mathbf{F}_v \odot \boldsymbol{\Phi}^{v}_k\right),
\end{align}
with $\boldsymbol{\Phi}^v_k$ learnable complex kernels and $\odot$ elementwise complex multiplication. The text branch is compressed analogously with its own filter bank and scoring head, yielding $\tilde{\mathbf{F}}_t$. Conditioning the mixture on the input spectrum, rather than fixing it, allows the effective filter to adapt per sample.

\paragraph{Cross-modal emphasize-and-suppress.} Each branch's compressed spectrum modulates the other. A source spectrum is pooled via learned-query attention, normalized, and mapped through channel- and band/position-specific linear heads to a bounded gate,
\begin{equation}
\sigma_b(x) = 0.1 + 0.8 \cdot \mathrm{sigmoid}(x),
\end{equation}
which keeps gate values in $[0.1, 0.9]$ so that a branch's frequency content can be strongly attenuated but never fully discarded or fully passed through. The question spectrum produces gate $\boldsymbol{\gamma}^v$, applied to the image spectrum; the image spectrum produces gate $\boldsymbol{\gamma}^t$, applied to the text spectrum:
\begin{equation}
\hat{\mathbf{F}}_v = \tilde{\mathbf{F}}_v \odot \boldsymbol{\gamma}^v(\tilde{\mathbf{F}}_t), \qquad
\hat{\mathbf{F}}_t = \tilde{\mathbf{F}}_t \odot \boldsymbol{\gamma}^t(\tilde{\mathbf{F}}_v).
\end{equation}

\paragraph{Inverse transform and fusion.} The gated spectra are mapped back via inverse FFTs and combined with their pre-transform inputs through a residual feed-forward sub-layer (pre-norm, with an internal residual connection, denoted $\mathrm{AddNorm}$):

\begin{align}
\mathbf{O}_v &= \mathrm{AddNorm}\!\left(\mathcal{F}^{-1}_{2D}(\hat{\mathbf{F}}_v) + \mathbf{V}_e\right), \\
\mathbf{O}_t &= \mathrm{AddNorm}\!\left(\mathcal{F}^{-1}_{1D}(\hat{\mathbf{F}}_t) + \mathbf{X}_t\right).
\end{align}
The branches are concatenated, projected, and normalized,
\begin{equation}
\mathbf{Z} = \mathrm{LayerNorm}\!\left(\mathrm{concat}(\mathbf{O}_v, \mathbf{O}_t)\, \mathbf{W}_o + \mathbf{b}_o\right),
\end{equation}
then mapped back into the language model's representation space, $\mathbf{Z}' = \mathrm{LayerNorm}(\mathbf{Z}\mathbf{W}_z + \mathbf{b}_z)$, and split into $\mathbf{Z}'_v \in \mathbb{R}^{N \times D}$ and $\mathbf{Z}'_t \in \mathbb{R}^{L_q \times D}$.

\subsection{Gated Multimodal Integration}

The frequency-processed streams are additively blended with the original visual and textual representations rather than replacing them, using two learnable scalar gates bounded to $[0.05, 0.95]$ by the same $\sigma_b$ function-style function:
\begin{align}
\bar{\mathbf{V}} &= \mathrm{LayerNorm}\!\left(\mathbf{V}_s + \alpha_v\, \mathbf{Z}'_v\right), \\
\bar{\mathbf{H}} &= \mathrm{LayerNorm}\!\left(\mathbf{H}_q + \alpha_t\, \mathbf{Z}'_t\right),
\end{align}
where $\alpha_v, \alpha_t$ are learned scalars, ensuring the spectral contribution is neither negligible nor dominant. A learnable source-type embedding is added to each stream so the decoder can distinguish visual from textual tokens:
\begin{equation}
\mathbf{V}^{\ast} = \bar{\mathbf{V}} + \mathbf{e}_v, \qquad \mathbf{H}^{\ast} = \bar{\mathbf{H}} + \mathbf{e}_t.
\end{equation}
The two streams are concatenated into a single fused encoder representation, with the attention mask formed by concatenating an all-ones visual mask with the question's padding mask:
\begin{equation}
\mathbf{E} = \mathrm{concat}(\mathbf{V}^{\ast}, \mathbf{H}^{\ast}) \in \mathbb{R}^{(N+L_q) \times D}.
\end{equation}

\subsection{Answer Generation and Auxiliary Objectives}

\paragraph{Generation.} $\mathbf{E}$ substitutes for the encoder output of a pretrained sequence-to-sequence decoder, which attends over $\mathbf{E}$ and autoregressively generates the answer:
\begin{equation}
\hat{a} = \arg\max_{a} \prod_{j=1}^{L_a} p_\theta\!\left(a^j \mid a^{<j}, \mathbf{E}\right).
\end{equation}
The decoder is supervised with teacher forcing using a label-smoothed cross-entropy loss $\mathcal{L}_{\mathrm{gen}}$ over non-padding positions.

\paragraph{Answer-type classification.} A masked mean pooling of $\mathbf{E}$ is passed through a linear head to predict the coarse answer type $y \in \{\text{yes/no}, \text{open}, \text{other}\}$:
\begin{equation}
\hat{y} = \mathrm{softmax}\!\left(\bar{\mathbf{e}}\,\mathbf{W}_y + \mathbf{b}_y\right), \qquad \bar{\mathbf{e}} = \frac{\sum_i m_i \mathbf{E}_i}{\sum_i m_i},
\end{equation}
with $m_i$ the attention mask at position $i$. This head is trained with a class-balanced focal loss~\citep{lin2017focal} $\mathcal{L}_{\mathrm{type}}$ to counteract answer-category imbalance.

\paragraph{Representation-level auxiliary losses.} Two auxiliary terms encourage the fused branches to carry answer-relevant, sample-specific information. A contrastive term $\mathcal{L}_{\mathrm{contrast}}$ pulls the correct (mean-pooled, $\ell_2$-normalized) image and text branch representations closer to a contextual encoding of the ground-truth answer than mismatched in-batch counterparts via a margin-based hinge, and additionally penalizes correct-pair similarity that falls below a fixed target, averaged over both branches. A diversity term $\mathcal{L}_{\mathrm{div}}$ penalizes high average pairwise cosine similarity between different samples' visual branch outputs within a batch, discouraging representational collapse.

\paragraph{Overall objective.} The full training loss is
\begin{equation}
\mathcal{L} = \mathcal{L}_{\mathrm{gen}} + \lambda_{\mathrm{type}}\,\mathcal{L}_{\mathrm{type}} + \lambda_{\mathrm{contrast}}\,\mathcal{L}_{\mathrm{contrast}} + \lambda_{\mathrm{div}}\,\mathcal{L}_{\mathrm{div}},
\end{equation}
with $\lambda_{\mathrm{type}}$, $\lambda_{\mathrm{contrast}}$, $\lambda_{\mathrm{div}} \geq 0$ non-negative weights, held fixed within a stage but reweighted at each stage transition to reflect that stage's training priorities; $\lambda_{\mathrm{div}}$ is held constant throughout.

%

\section{Experiments}

\subsection{Experimental Setup}

\paragraph{Datasets.}
The proposed model is pretrained on PMC-VQA ~\citep{zhang2023pmcvqa} and subsequently fine-tuned on two publicly available medical VQA benchmarks: VQA-RAD and SLAKE.VQA-RAD contains 1,793 official training QA pairs, from which we carve a 269-sample validation set via an image-level 85/15 split (ensuring no image appears in both splits), leaving 1,524 samples for training; the official 451-sample test set is used unmodified for evaluationSLAKE consists of 4,165 training, 1,053 validation, and 1,061 testing samples after restricting to the English-language subset, which we use throughout for consistency with the evaluation protocol. Following standard practice, all input images are resized to \(224 \times 224\).

\paragraph{Implementation Details.}
Our framework employs BiomedCLIP~\citep{zhang2023biomedclip} as the frozen visual encoder and BioBART-v2~\citep{yuan2022biobart} as the text encoder-decoder. The model is optimized using AdamW~\citep{loshchilov2019adamw} with mixed-precision training and a linear-warmup, cosine-decay learning-rate schedule. Pretraining is performed on PMC-VQA for 20 epochs, followed by fine-tuning on VQA-RAD for 10 epochs and SLAKE for 15 epochs, using a base learning rate of $5\times10^{-6}$ (discriminatively reduced to $0.1\times$ for the pretrained text encoder and, once unfrozen, the vision backbone; annealed to $1\times10^{-6}$ in the final fine-tuning stage). All experiments were conducted on a Google Colab G4 GPU (NVIDIA RTX PRO 6000 Blackwell Server Edition) using PyTorch 2.0 with AMP.

\paragraph{Evaluation Metrics.}
Following prior medical VQA studies, performance is evaluated using Exact Match (EM) and Token-level F1. Results are reported on the official test sets, with separate evaluation on open-ended and closed-ended question categories whenever applicable.

\subsection{Baselines}
We compare against established Med-VQA methods evaluated under the same
Exact Match protocol on VQA-RAD and SLAKE: M3AE~\cite{chen2022m3ae},
M2I2~\cite{li2023m2i2} (262.15M params, closest in scale to our 260M model),
MUMC~\cite{li2023mumc} (211.06M params), and PeFoMed~\cite{he2024pefomed}.
All baseline scores are taken directly from their original publications.
Our results are averaged over 3 seeds (42, 123, 2024) on the official test splits.

\section{Results and Analysis}
\begin{table*}[t]
\centering
\caption{Accuracy (Exact Match, \%) on VQA-RAD and SLAKE test sets.}
\label{tab:baselines}
\begin{tabular}{lccc ccc}
\toprule
\multirow{2}{*}{Method} & \multicolumn{3}{c}{VQA-RAD} & \multicolumn{3}{c}{SLAKE} \\
\cmidrule(lr){2-4} \cmidrule(lr){5-7}
 & Open & Closed & Overall & Open & Closed & Overall \\
\midrule
M3AE            & 67.2 & 83.5 & 77.0 & 80.3 & 87.8 & 83.2 \\
M2I2            & 66.5 & 83.5 & 76.8 & 74.7 & 91.1 & 81.2 \\
MUMC            & 71.5 & 84.2 & 79.2 & 81.5 & 91.1 & 84.9 \\
PeFoMed         & 62.6 & 87.1 & 77.4 & 77.8 & 88.7 & 82.1 \\
\midrule
Ours & 19.8 & 56.4 & 40.2 $\pm$ 0.6 & 68.3 & 70.7 & 69.2 $\pm$ 1.1 \\
\bottomrule
\end{tabular}
\end{table*}

Table~\ref{tab:baselines} presents the quantitative comparison of the proposed method with representative medical VQA approaches on the VQA-RAD and SLAKE benchmarks. Our model achieves an overall Exact Match (EM) score of 40.21\% on VQA-RAD and 69.21\% on SLAKE, with standard deviations of 0.56\% and 1.10\%, respectively, across three random seeds. The relatively small variance indicates that the proposed training strategy produces stable and reproducible results. In addition, the proposed model achieves Token F1 scores of 44.01\% on VQA-RAD (Open 28.4, Closed 56.4) and 72.90\% on SLAKE (Open 74.7, Closed 70.7), suggesting that many generated answers are semantically relevant even when they do not exactly match the reference answer.

Compared with recent Med-VQA methods, including M3AE, M2I2, MUMC, and PeFoMed, the proposed framework achieves lower overall benchmark performance. These methods benefit from stronger multimodal pretraining strategies and specialized architectures designed specifically for medical visual question answering. Nevertheless, our objective is not to replace these systems directly, but rather to investigate whether frequency-domain multimodal fusion can improve generative medical VQA. Despite its relatively lightweight architecture based on BiomedCLIP and BioBART, the proposed model demonstrates competitive performance on the larger SLAKE benchmark and provides a practical foundation for evaluating the effectiveness of spectral feature fusion.

A clear difference is observed between the two datasets. The proposed method performs considerably better on SLAKE than on VQA-RAD, particularly for open-ended questions. A possible contributing factor is the substantially larger training set provided by SLAKE, which offers more supervision for learning robust multimodal representations. In contrast, VQA-RAD contains fewer training examples and a wider variety of answer expressions, making exact-match evaluation substantially more challenging for generative models. This trend is also reflected in the Token F1 scores, which consistently exceed Exact Match, indicating that the generated responses often contain partially correct medical information despite failing strict string-level matching.

\subsection{Ablation Study}

To evaluate the contribution of the proposed Frequency Spectrum Representation and Fusion Unit (FSRU), we compare the complete Q-FSRU framework with a variant in which the FSRU module is removed while keeping all other components and the training protocol unchanged. The results are summarized in Table~\ref{tab:ablation}.

\begin{table}[t]
\centering
\small
\caption{Ablation study on the FSRU module. Results are mean $\pm$ std over three seeds.}
\label{tab:ablation}
\begin{tabular}{lcc}
\toprule
Configuration & VQA-RAD EM (\%) & SLAKE EM (\%) \\
\midrule
w/o FSRU & 36.66 $\pm$ 0.78 & 66.26 $\pm$ 0.09 \\
\textbf{Ours (Full Model)} & \textbf{40.21 $\pm$ 0.56} & \textbf{69.21 $\pm$ 1.10} \\
\bottomrule
\end{tabular}
\end{table}

The proposed model consistently outperforms the variant without the FSRU module on both VQA-RAD and SLAKE. Specifically, incorporating the proposed frequency-domain fusion module improves Exact Match (EM) from 36.66\% to 40.21\% on VQA-RAD and from 66.26\% to 69.21\% on SLAKE. These improvements indicate that the FSRU module enhances multimodal feature fusion and contributes to more accurate answer generation.

The consistent performance gains across both benchmarks demonstrate the effectiveness of the proposed frequency-domain fusion strategy. Although the proposed model does not achieve state-of-the-art performance, the ablation study shows that the FSRU module makes a positive contribution to the overall framework. These findings suggest that frequency-domain multimodal fusion is a promising direction for generative medical visual question answering and could be further improved through larger vision-language models and more extensive medical pretraining.

\subsection{Qualitative Analysis}
\begin{figure}[b]
    \centering
    \includegraphics[width=0.8\linewidth]{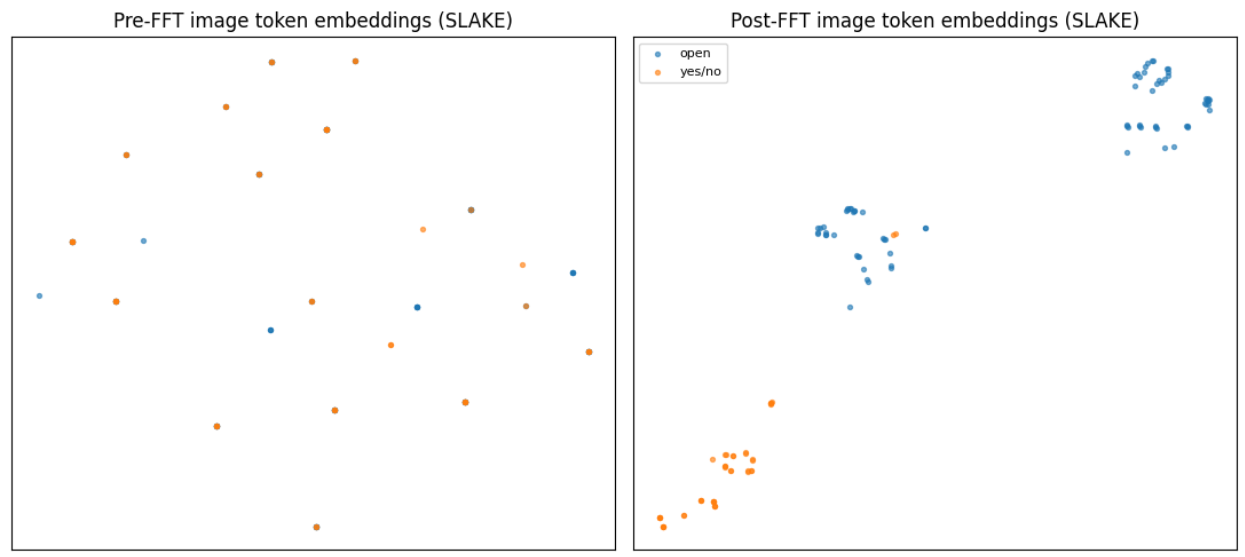}
    \caption{t-SNE~\citep{vandermaaten2008tsne} visualization of the image token embeddings before and after the proposed FSRU module on the SLAKE validation set.}
    \label{fig:fsru_tsne.png}
\end{figure}

Figure~\ref{fig:fsru_tsne.png} presents a t-SNE visualization of the image token embeddings before and after the proposed FSRU module on the SLAKE validation set. Before frequency-domain fusion, the embeddings are widely scattered with noticeable overlap between different question categories. After applying FSRU, the embeddings become more compact and better separated, indicating that the proposed frequency-spectrum fusion learns a more structured and discriminative feature representation. This improved feature organization is consistent with the quantitative performance gains achieved by the proposed model.

\subsection{Case study}
\begin{figure}[ht]
    \centering
    \includegraphics[width=0.8\linewidth]{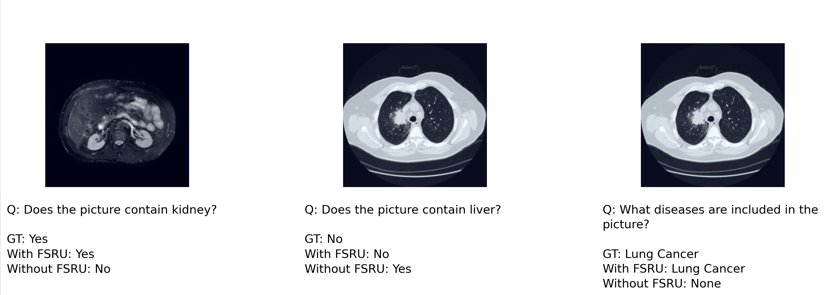}
    \caption{Representative case studies on the SLAKE validation set comparing the proposed model with and w/o FSRU. The proposed model correctly answers several medical VQA questions that are incorrectly or incompletely predicted by the ablated model, demonstrating the effectiveness of frequency-domain multimodal fusion.}
\label{fig:case_studies}
\end{figure}

Figure~\ref{fig:case_studies} shows representative predictions with and without FSRU. Incorporating FSRU correctly identifies anatomical structures, diseases, and abnormality locations that the ablated model misses, illustrating the benefit of frequency-domain fusion for clinical reasoning.

\begin{figure}[ht]
    \centering
    \includegraphics[width=0.8\linewidth]{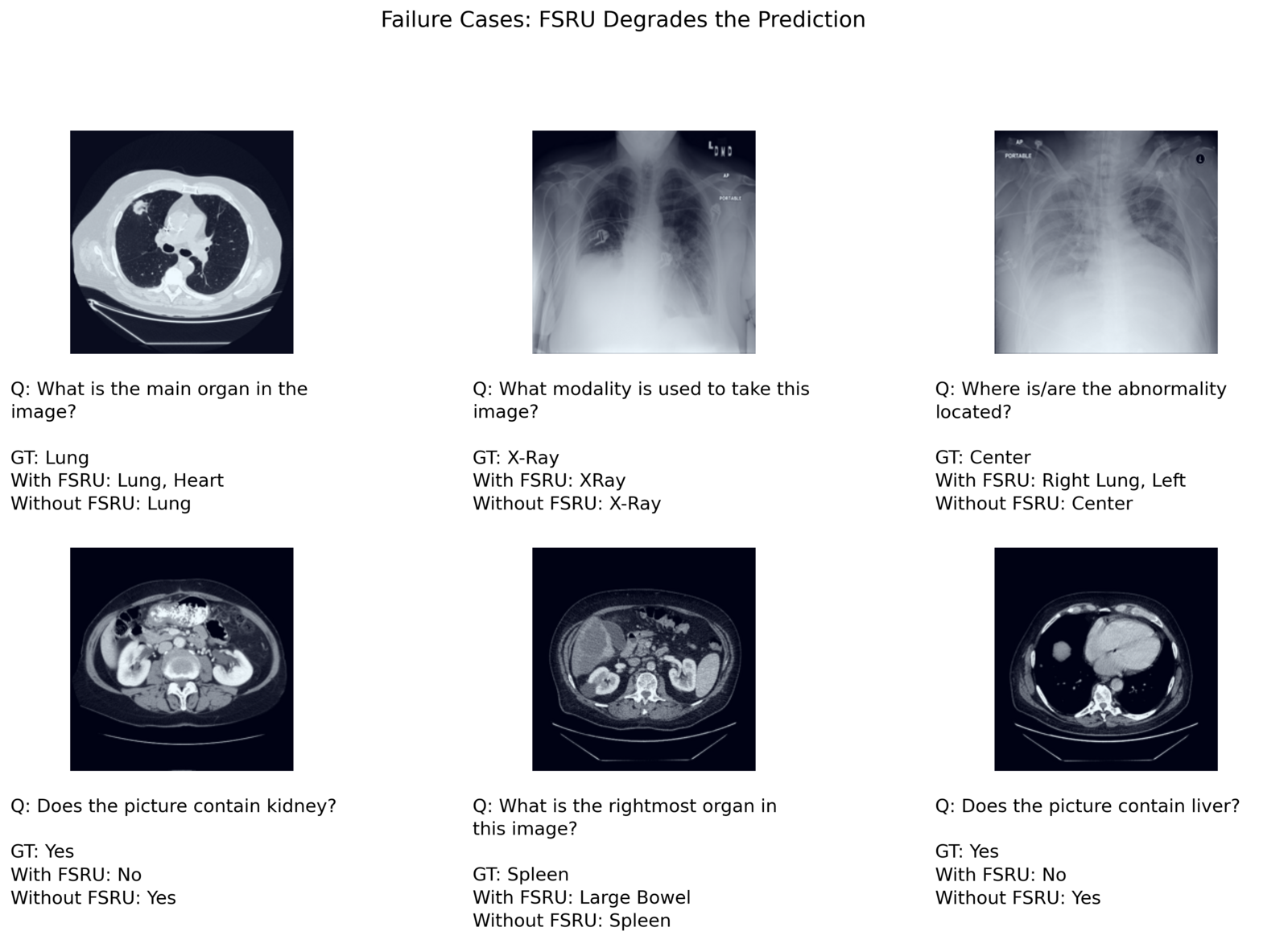}
    \caption{Representative failure cases where FSRU degrades predictions relative to the w/o-FSRU baseline on the SLAKE validation set. FSRU occasionally flips correct closed-ended (yes/no) organ-presence answers, and on abdominal CT slices with several adjacent organs, produces over-inclusive or malformed multi-token spans (e.g., predicting two organs, or an incoherent location phrase) where the baseline gives a single correct answer.}
    \label{fig:fsru_failure_cases}
\end{figure}

Despite these gains, FSRU is not uniformly beneficial. Figure~\ref{fig:fsru_failure_cases} shows representative cases where it degrades predictions relative to the w/o-FSRU baseline. On closed-ended organ-presence questions, FSRU occasionally flips a correct binary answer, and on abdominal CT slices containing several adjacent organs, it produces over-inclusive or malformed multi-token spans, such as naming two organs where only one was asked about or generating an incoherent locational phrase, whereas the baseline produces a single well-formed answer. These failures are concentrated in denser, multi-structure CT images, suggesting that cross-modal gating can occasionally over-emphasize competing high-frequency content when several anatomically similar structures share a spectral neighborhood, an effect the question-conditioning does not fully suppress.

\subsection{Conclusion}
We presented a dual-branch frequency-domain fusion framework for medical VQA, combining BiomedCLIP and BioBART with a question-guided Frequency Spectral Representation Unit (FSRU). The ablation study confirms FSRU consistently improves answer prediction over the non-frequency baseline, though results do not surpass current state-of-the-art methods. Our failure case analysis further indicates that these gains are not uniform: FSRU is most beneficial on open-ended, texture-sensitive questions, but can occasionally degrade closed-ended organ-presence answers and produce over-inclusive multi-token spans on visually dense, multi-organ CT slices. Future work will explore larger vision-language backbones, additional imaging datasets, and gating mechanisms with tighter per-token control to mitigate these failure modes. 

\bibliography{aaai2027}
\end{document}